\documentclass[11pt,a4paper]{article}
\usepackage[hyperref]{acl2021}
\usepackage{times}
\usepackage{latexsym}

\usepackage{verbatim}

\usepackage{microtype}
\usepackage{listings}

\aclfinalcopy 

\usepackage{paralist}
\usepackage{xr}

\makeatletter
\newcommand*{\addFileDependency}[1]{
  \typeout{(#1)}
  \@addtofilelist{#1}
  \IfFileExists{#1}{}{\typeout{No file #1.}}
}
\makeatother

\usepackage{tabularx,lipsum,environ,amsmath,amssymb}

\makeatletter

\newcolumntype{\expand}{}
\long\@namedef{NC@rewrite@\string\expand}{\expandafter\NC@find}

\NewEnviron{problem}[2][]{%
  \def\problem@arg{#1}%
  \def\problem@framed{framed}%
  \def\problem@lined{lined}%
  \def\problem@doublelined{doublelined}%
  \ifx\problem@arg\@empty%
    \def\problem@hline{}%
  \else%
    \ifx\problem@arg\problem@doublelined%
      \def\problem@hline{\hline\hline}%
    \else%
      \def\problem@hline{\hline}%
    \fi%
  \fi%
  \ifx\problem@arg\problem@framed%
    \def\problem@tablelayout{|>{\bfseries}lX|c}%
    \def\problem@title{\multicolumn{2}{|l|}{%
        \raisebox{-\fboxsep}{\textsc{\normalsize #2}}%
      }}%
  \else
    \def\problem@tablelayout{>{\bfseries}lXc}%
    \def\problem@title{\multicolumn{2}{l}{%
        \raisebox{-\fboxsep}{\textsc{\normalsize #2}}%
      }}%
  \fi%
  \bigskip\par\noindent%
  \begin{tabularx}{\textwidth}{\expand\problem@tablelayout}%
    \problem@hline%
    \problem@title\\[2\fboxsep]%
    \BODY\\\problem@hline%
  \end{tabularx}%
  \medskip\par%
}
\makeatother

\usepackage{url}

\usepackage{footnote}
\usepackage{xspace}
\usepackage{tikz}
\usepackage{tikz-qtree-compat}
\usetikzlibrary{shapes, fit, arrows.meta}
\usepackage{lingmacros}
\usepackage{colortbl} 
\usepackage{tabularx}
\usepackage{flafter} 
\usepackage{float}
\usepackage{graphicx}
\usepackage[skins]{tcolorbox}
\tikzset{>=latex}
\usepackage{footmisc}
\usepackage{booktabs}

\newsavebox{\fmbox}

\usepackage[color=red!5]{todonotes}
\let\todonote\todo

\newcommand{\inlinetext}[1]{``\textit{#1}''}

\usepackage{xcolor}
\usepackage[skins]{tcolorbox}

\definecolor{darkblue}{rgb}{0, 0, 0.5}
\hypersetup{colorlinks=true,citecolor=darkblue, linkcolor=darkblue, urlcolor=darkblue}
\definecolor{darkolivegreen}{rgb}{0.33, 0.42, 0.18}
\definecolor{maroon(html/css)}{rgb}{0.5, 0.0, 0.0}
\definecolor{darkslategray}{rgb}{0.18, 0.31, 0.31}
\definecolor{darksienna}{rgb}{0.24, 0.08, 0.08}
\definecolor{darkscarlet}{rgb}{0.34, 0.01, 0.1}
\definecolor{airforceblue}{rgb}{0.36, 0.54, 0.66}
\definecolor{beaublue}{rgb}{0.74, 0.83, 0.9}
\definecolor{applegreen}{rgb}{0.55, 0.71, 0.0}
\definecolor{amber(sae/ece)}{rgb}{1.0, 0.49, 0.0}
\definecolor{ao(english)}{rgb}{0.0, 0.5, 0.0}

\usepackage{soul}
\sethlcolor{beaublue}

\usepackage{float}
\usepackage{graphicx}
\usepackage{wrapfig}
\usepackage{floatflt}
\input{insbox.tex}

\usepackage{multirow}
\usepackage{xspace} 
\usepackage{fancyvrb} 
\usepackage{caption} 
\usepackage{xstring}
\usepackage{soul}
\usepackage[normalem]{ulem} 
\usepackage{paralist} 
\usepackage{algpseudocode,algorithm} 
\usepackage{tikz}
\usepackage{tikz-qtree-compat}
\usetikzlibrary{shapes, fit, arrows.meta}
\usepackage{lingmacros}
\usepackage{colortbl} 
\usepackage{tabularx}
\usepackage{flafter} 

\usepackage{adjustbox}

\algnewcommand{\LineComment}[1]{\State \(\triangleright\) #1} 

\algrenewcommand\alglinenumber[1]{\scriptsize #1:} 

\tikzset{>=latex}

\renewcommand{\todo}[1]{\todonote[size=\tiny]{#1}{\textcolor{red}{(TODO: #1)}}}
\newcommand{\annot}[2][]{\todonote[size=\tiny]{#1}{\textcolor{red}{(#2)}}}

\usepackage{fancyvrb}

\usepackage{url}

\title{Context-Preserving Text Simplification} 

\author{Christina Niklaus\textsuperscript{1}\textsuperscript{3}, Matthias Cetto\textsuperscript{1}, Andr\'{e} Freitas\textsuperscript{2}, \and Siegfried Handschuh\textsuperscript{1}\textsuperscript{3} \\
  \textsuperscript{1} University of St.Gallen\\
  {\small{{\tt \{christina.niklaus, matthias.cetto, siegfried.handschuh\}{\tt @unisg.ch}}}}\\
  \textsuperscript{2} University of Manchester\\
  {\small{{\tt andre.freitas@manchester.ac.uk}}}\\
  \textsuperscript{3} University of Passau\\
  {\small{{\tt \{christina.niklaus, siegfried.handschuh\}{\tt @uni-passau.de}}}}
\\}

\date{}

\begin{document}
\maketitle
\begin{abstract}
We present a context-preserving text simplification (TS) approach that recursively splits and rephrases complex English sentences into a semantic hierarchy of simplified sentences. Using a set of linguistically principled transformation patterns, input sentences are converted into a 
 hierarchical representation in the form of core sentences and accompanying contexts that are linked via rhetorical relations. Hence, as opposed to previously proposed sentence splitting approaches, which commonly do not take into account discourse-level aspects, our TS approach preserves the semantic relationship of the decomposed constituents in the output. 
  A comparative analysis with the annotations contained in the RST-DT shows that we are able to capture the contextual hierarchy between the split sentences with a precision of 89\% and reach an average precision of 69\% for the classification of the rhetorical relations that hold between them. 
\end{abstract}

\section{Introduction}


Sentences that present a complex linguistic structure can be \textit{hard to comprehend by human readers}, as well as \textit{difficult to analyze by semantic applications}.  
Identifying grammatical complexities in a sentence and transforming them into simpler structures 
is the goal of syntactic TS. One of the major types of operations that are used to perform this rewriting step is \textit{sentence splitting}: it divides a sentence into several shorter components, with each of them presenting a simpler and more regular structure that is easier to process by both humans \cite{Siddharthan2014,saggion-2015-simplext,ferres-2016-yats} and machines \cite{stajner-popovic-2016-text,stajner-popovic-2018-improving,saha-mausam-2018-open}. 

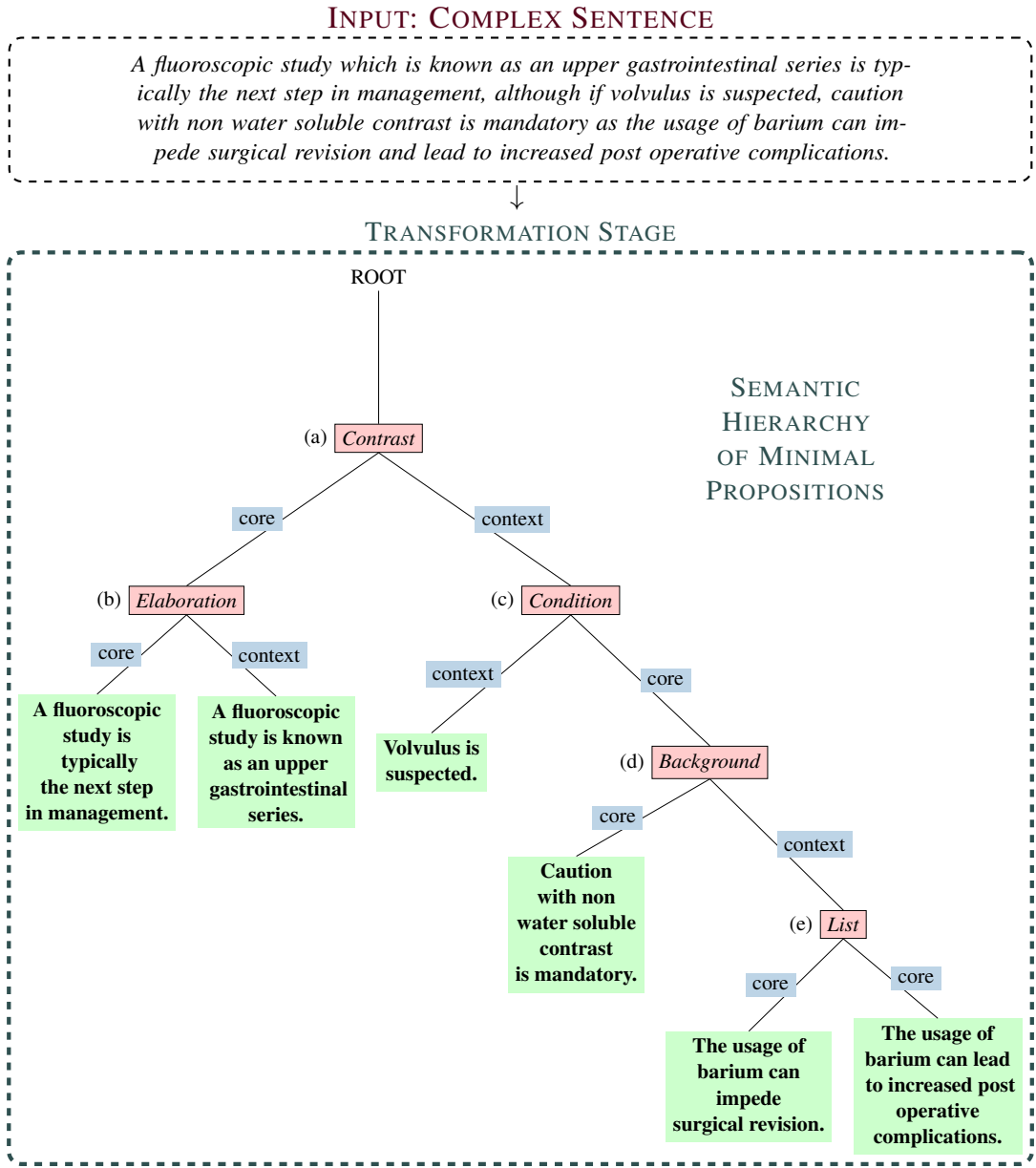
\begin{figure*}[!htb]
\centering

\begin{minipage}{\linewidth}
\centering
\begin{tikzpicture}[scale=0.85, every node/.style={align=center, transform shape}]
\node (1)[rectangle, solid, text width=\linewidth, rounded corners=5pt]{
\textit{{A fluoroscopic study which is known as an upper gastrointestinal series is typically the next step in management, although if volvulus is suspected, caution with non water soluble contrast is mandatory as the usage of barium can impede surgical revision and lead to increased post operative complications.}}
};

\node (2)[draw,dashed, thick,fit=(1), label=above:\Large{\textsc{\textcolor{darkscarlet}{Input: Complex Sentence}}}, rounded corners=5pt] {};

\end{tikzpicture}
\end{minipage}%

\vspace{0.0cm}
$\downarrow$
\vspace{0.0cm}

\begin{minipage}{\textwidth}
\centering
\begin{tikzpicture}[scale=0.75, level distance=3cm, sibling distance=0.4cm, every tree node/.style={align=center, transform shape}]
\Tree [
 .\node(z){ROOT};
    \edge node[midway, left] {};
      [.\node (y)[style={draw,rectangle}, label=left:(a), fill=red!20] {\textit{Contrast}};
            \edge node[midway, left, fill=beaublue] {core}; [
                .\node [style={draw,rectangle}, label=left:(b), fill=red!20] {\textit{Elaboration}};
                    \edge node[midway, left, fill=beaublue] {core}; [.\node(a) [fill=green!20]{\textbf{A fluoroscopic}\\ \textbf{study is}\\\textbf{ typically}\\ \textbf{the next step}\\ \textbf{in management.}};]
                    \edge node[midway, right, fill=beaublue] {context}; 
                    [.\node(b) [fill=green!20]{\textbf{A fluoroscopic} \\ \textbf{study is known}\\ \textbf{as an upper}\\\textbf{ gastrointestinal}\\ \textbf{series.}};]
            ]
            \edge node[midway, right, fill=beaublue] {context}; [
                .\node [style={draw,rectangle}, label=left:(c), fill=red!20] {\textit{Condition}};
                     \edge node[midway, left, fill=beaublue] {context}; [.\node(c) [fill=green!20]{\textbf{Volvulus is}\\ \textbf{suspected.}};]
                    \edge node[midway, right, fill=beaublue] {core};[.\node [style={draw,rectangle}, label=left:(d), fill=red!20] {\textit{Background}};
                        \edge node[midway, left, fill=beaublue] {core}; [.\node(d) [fill=green!20]{\textbf{Caution}\\ \textbf{with non}\\ \textbf{water soluble}\\ \textbf{contrast}\\ \textbf{is mandatory.}};]
                        \edge node[midway, right, fill=beaublue] {context}; [.\node [style={draw,rectangle}, label=left:(e), fill=red!20] {\textit{List}};
                            \edge node[midway, left, fill=beaublue] {core}; [.\node(e) [fill=green!20]{\textbf{The usage of}\\ \textbf{barium can}\\ \textbf{impede}\\ \textbf{surgical revision.}};]
                            \edge node[midway, right, fill=beaublue] {core}; [.\node(f) [fill=green!20] {\textbf{The usage of}\\ \textbf{barium can lead}\\\textbf{ to increased post}\\ \textbf{operative}\\ \textbf{complications.}};]
                        ]
                    ]
            ]
            ]
        ]
]

\node (9)[draw,dashed, ultra thick, color=darkslategray, fit=(a) (b) (c) (d) (e) (f) (z), label=above:\textsc{\textcolor{darkslategray}{Transformation Stage}}, rounded corners=5pt] {};


\node (5)[text width = 3cm, align=center, right =of y, xshift=2.5cm]{\textcolor{darkslategray}{\textsc{Semantic Hierarchy of Minimal Propositions}}};



\end{tikzpicture}

\end{minipage}

\caption{A complex sentence is transformed into a semantic hierarchy of simplified sentences in the form of minimal, self-contained propositions that are linked to each other via rhetorical relations. The output presents a regular, fine-grained structure that preserves the context of the input in the form of hierarchically ordered and semantically interconnected sentences.}
\label{intro_example1}
\end{figure*}

We propose a sentence splitting approach that can be used as a preprocessing step to generate an intermediate representation. The objective is to facilitate and improve the performance of downstream 
tasks whose predictive quality deteriorates with sentence length and complexity. 
Our approach aims to \textbf{break down a complex sentence into a set of minimal propositions}, i.e. a sequence of sound, self-contained utterances with a simple and regular structure. Each of them presents a minimal unit of coherent information and cannot be further decomposed into meaningful propositions.

However, any sound and coherent text is not simply a loose arrangement of self-contained units, but rather a logical structure of utterances that are semantically connected \cite{siddharthan2014survey}. Consequently, when carrying out syntactic TS operations without considering discourse implications, the rewriting may easily result in a disconnected sequence of simplified sentences, making the text harder to interpret. 
The vast majority of existing structural TS approaches though do not take into account discourse-level aspects. Therefore, they are prone to producing a set of incoherent utterances where important contextual information is lost. 
Thus, in order to \textbf{preserve the coherence structure} of the input 
we propose a context-preserving TS approach. 
It establishes a semantic hierarchy between the split components by (1) setting up a contextual hierarchy and (2) classifying the semantic relationship that holds between them (see Figure \ref{intro_example1}).


To the best of our knowledge, this is the first time that syntactically complex sentences are \textit{split and rephrased within the semantic context} in which they occur. 
Our framework differs from previously proposed approaches by using a linguistically grounded
transformation stage that applies clausal and phrasal disembedding mechanisms to transform sentences into shorter utterances with a more regular structure.
By using a recursive top-down approach, it generates a novel \textit{hierarchical representation} between those units, capturing both their semantic context and relations to other units in the form of rhetorical relations. By taking advantage of the resulting fine-grained representation, 
the complexity of downstream tasks may be reduced, thus improving their performance. In addition, by incorporating the semantic context of the source sentences, our proposed representation preserves important contextual information that is needed to maintain the coherence structure of the input, allowing for a proper interpretation of complex assertions.

In summary, we make the following contributions: (i) We conduct a systematic analysis of syntactic phenomena involved in complex sentence constructions whose findings are materialized in a small set of 35 hand-crafted transformation rules. (ii) We propose a context-preserving syntactic TS approach which transforms complex sentences 
into a semantic hierarchy of minimal propositions.  
(iii) The proposed method is linguistically grounded and does not require any training data. 
(iv) As a proof of concept, we develop a reference implementation. (v) We perform a comprehensive empirical evaluation, demonstrating that we reach state-of-the-art performance in the classification of both the hierarchical order and the semantic relationship that hold between the split sentences.

\section{Context-Preserving Sentence Splitting}

We present \textsc{DisSim}, a context-preserving TS approach that creates a semantic hierarchy of simplified sentences.\footnote{The source code of our framework is publicly available for download under the following link: \url{https://github.com/Lambda-3/DiscourseSimplification}.} It takes a sentence as input and performs a recursive transformation stage that is based upon a small set of 35 hand-crafted 
rules. 

\subsection{Transformation Patterns}

\begin{table*}[!htb]
\small
\centering
  \begin{tabular}{ l }
    \toprule
   {ROOT $<<:$ (S $<$ (NP $\$..$ (VP $<+$(VP) \underline{(\textbf{SBAR} $<,$ (\textit{IN} $\$+$ (\fbox{S $<$ (NP $\$..$ VP}))))})))} \\ \bottomrule

    \end{tabular}
  
  \caption{Example of a transformation pattern (for decomposing adverbial clauses). 
  They are specified in terms of Tregex patterns \cite{levy-andrew-2006-tregex}. 
  A boxed pattern represents the part of a sentence that is extracted from the input and transformed into a new stand-alone sentence. A pattern in bold is deleted from the source. The underlined part is labelled as a context sentence, while the remaining part represents core information. The italic pattern is used as a cue phrase for identifying the rhetorical relation that holds between the decomposed spans.}
  \label{examplePatterns}

\end{table*}

In the development of the transformation patterns, we followed a principled and systematic procedure, with the goal of eliciting a universal set of transformation rules. They were heuristically determined in a rule-engineering process that was carried out on the basis of an in-depth study of the literature on syntactic sentence simplification, e.g.  \cite{siddharthan2006syntactic,siddharthan2014survey,siddharthan2002architecture,Siddharthan2014,evansorasan2019,heilman-2010-extracting,Saggion2018book,mallinson2019controllable,ferres-2016-yats}.
Next, we performed a thorough linguistic analysis of the syntactic phenomena that need to be tackled in the sentence splitting task. Details on the underlying linguistic principles, supporting the systemacity and universality of the developed transformation patterns, can be found in Section \ref{app:pattern_development} in the appendix. 


The transformation patterns encode syntactic and lexical features that can be derived from a sentence's phrase structure.
Each rule specifies (1) how to \textit{split up and rephrase} the input into structurally simplified sentences and (2) how to \textit{set up a contextual hierarchy} between the split components and how to \textit{identify the semantic relationship} that holds between those elements. An example of a transformation rule is provided in Table \ref{examplePatterns}.\footnote{For reproducibility purposes, the full set of patterns can be found online: \url{https://github.com/Lambda-3/DiscourseSimplification/tree/master/supplemental_material}.} 

\subsection{Data Model: Linked Proposition Tree}

\begin{figure}[!htb]
\centering
\begin{tikzpicture}[scale=0.78, transform shape]
\node(1)[ellipse, draw, fill=green!20, minimum width = 3cm, align = center, minimum height = 1cm, label=below:{}, xshift=-3cm] {$prop \in PROP$};
\node(2)[ellipse, draw, fill=green!20, minimum width = 3cm, align = center, minimum height = 1cm, xshift = 3cm, label=below:{}] {$prop \in PROP$};
\node(3)[draw, fill=red!20, minimum width = 2cm, align = center, minimum height = 1cm, xshift = 0cm, yshift=2cm] {$rel \in REL$};


\draw[-, very thick] (1) -- node[above left] {\colorbox{beaublue}{\small{$c \in CL$}}} (3);
\draw[-, very thick] (2) -- node[above right] {\colorbox{beaublue}{\small{$c \in CL$}}} (3);

\end{tikzpicture}

\caption{Basic structure of a linked proposition tree $LPT$. It represents the data model of the semantic hierarchy of minimal propositions.}
\label{fig:data_model}
\end{figure}
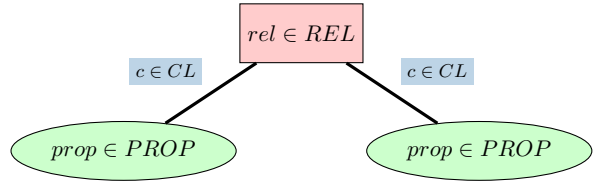

The transformation algorithm takes a complex sentence as input and recursively transforms it into a semantic hierarchy of minimal propositions. The output is represented as a linked proposition tree. Its basic structure is depicted in Figure \ref{fig:data_model}. A linked proposition tree is a labeled binary tree 
$LPT=(V, E)$.

Let $V\in \{REL, PROP\}$ be the set of nodes, where $PROP$ is the set of leaf nodes denoting the set of \colorbox{green!20}{minimal propositions}. A $prop \in PROP$ is a triple $(s, v, o) \in CT$, where $CT=\{$\textit{SV, SVA, SVC, SVO, SVOO, SVOA, SVOC}$\}$ represents the set of clause types \cite{delcorro-2013-clausie}. Hence, $s \in S$ denotes a subject, $v \in V$ a verb and $o \in \{O, A, C, OO, OA, OA, \emptyset \}$ a direct or indirect object, adverbial or complement (or a combination thereof). Accordingly, a minimal proposition $prop \in PROP$ is a simple sentence\footnote{A simple sentence comprises exactly one clause.} that is reduced to its clause type.\footnote{In addition, a specified set of phrasal elements were extracted. The interested reader may refer to Section \ref{app:pattern_development} in the appendix for more details.}
Thus, it represents a minimal unit of coherent information where all optional constituents are discarded, resulting in an utterance that expresses a single event consisting of a predicate and its core arguments. 


Furthermore, let $REL=\{$\textit{Contrast, List, Disjunction, Cause, Result, Temporal, Background, Condition, Elaboration, Explanation, Spatial, Attribution, Unknown}$\}$ be the set of \colorbox{red!20}{rhetorical relations}, comprising the set of inner nodes. A $rel \in REL$ represents the semantic relationship that holds between its child nodes. It reflects the semantic context of the associated propositions $prop \in PROP$. In that way, the coherence structure of the input is preserved.

Finally, let $E \in CL $, with $CL \in \{$\textit{core, context}$\}$, be the set of \colorbox{beaublue}{constituency labels}. A $c \in CL$ represents a labeled edge that connects two nodes $V \in LPT$. It enables the distinction between core information and less relevant contextual information. In that way, hierarchical structures between the split propositions $prop \in PROP$ are captured. Figure \ref{intro_example1} shows a concrete example of the linked proposition tree that is generated by our TS approach on an example sentence.





\subsection{Transformation Algorithm}

The transformation algorithm of our 
approach is given in Algorithm \ref{alg:transformation}. It takes a sentence as input and applies the  
transformation patterns to recursively transform it into a semantic hierarchy of minimal propositions, represented as an $LPT$. 


%

\begin{algorithm}[!htb]
\scriptsize
\caption{Transform into Semantic Hierarchy}
\label{alg:transformation}
\begin{algorithmic}[1]
    \Require{complex source sentence $str$}
    \Ensure{linked proposition tree $tree$}
    \Statex
    \Function{Initialize}{$str$}
        \State $new\_leaves$ $\gets$ source sentence $str$
        \State $new\_node \gets$ create a new parent node for $new\_leaves$
        \State $new\_node.labels$ $\gets$ None
        \State $new\_node.rel$ $\gets$ ROOT
        \State linked proposition tree $tree$ $\gets$ initialize with $new\_node$
        \State \Return $tree$
    \EndFunction
    \Statex
    \Procedure{TraverseTree}{$tree$}
      \LineComment{Process leaves (i.e. propositions) from left to right}
      \For {$leaf$ in $tree.leaves$}
        \LineComment{Check transformation rules in fixed order}
        \For {$rule$ in $TRANSFORM\_RULES$}
       \If{$match$} 
          \LineComment{\colorbox{green!20}{\textbf{(a) Sentence splitting}}} 
          \State 
          \colorbox{green!20}{$simplified\_propositions \gets$ decompose $leaf$ into a }
          \State \colorbox{green!20}{set of simplified propositions}
          \State \colorbox{green!20}{$new\_leaves \gets$ convert $simplified\_propositions$}
          \State \colorbox{green!20}{into leaf nodes}
          \LineComment{\colorbox{beaublue}{\textbf{(b) Constituency Type Classification}}} 
          \State 
          \colorbox{beaublue}{$new\_node \gets$ create a new parent node for $new\_leaves$}
          \State \colorbox{beaublue}{$new\_node.labels \gets$ link each leaf in $new\_leaves$ to}
          \State \colorbox{beaublue}{ $new\_node$ and label each edge with the leaf's constituency} 
          \State \colorbox{beaublue}{ type $c \in CL$}
          
          
          \LineComment{\colorbox{red!20}{\textbf{(c) Rhetorical Relation Identification}}}
          \State \colorbox{red!20}{$cue\_phrase \gets$ extract cue phrase from $leaf.parse\_tree$}
          \State \colorbox{red!20}{$new\_node.rel \in REL \gets$ match $cue\_phrase$ against a}
          \State \colorbox{red!20}{ predefined set of rhetorical cue words}
          
          \LineComment{Update Tree}
          \State $tree.replace(leaf, new\_node)$
          \LineComment{Recursion}
          \State \Call{TraverseTree}{$tree$}
         
        \EndIf
   \EndFor
   \EndFor
   \State \Return $tree$
    \EndProcedure
\end{algorithmic}
\end{algorithm}

\paragraph{Initialization} In the initialization step (1-7), the linked proposition tree $LPT$ is instantiated. At this point, it consists of the source sentence, which is represented as a single leaf node, and an unlabeled edge to the root node.

\paragraph{Tree Traversal} Next, the linked proposition tree is recursively traversed, splitting up the input in a top-down approach (9). Starting from the root node, the leaves are processed in depth-first order. For every leaf (11), we check if its phrasal parse tree matches one of the transformation patterns (13). The rules are applied in a fixed order that was empirically determined.\footnote{See Section \ref{app:application_order} in the appendix.} 
The first pattern that matches the proposition's parse tree will be employed (14). For instance, the first rule that matches the source sentence from Figure \ref{intro_example1} is the pattern depicted in Table \ref{examplePatterns}.

\begin{figure}[!htb]
\begin{center}
\begin{tikzpicture}[scale=0.66, level distance=2.7cm, sibling distance=0.5cm, every tree node/.style={align=center}]
\Tree [.\node[style={draw,rectangle}, fill=red!20] {\textbf{(c)} \inlinetext{although} $\rightarrow$ \textit{Contrast}}; 
  \edge node[midway, left, fill=beaublue] {\textbf{(b)} core}; {\colorbox{green!20}{\textbf{(a)}} \\\colorbox{green!20}{A fluoroscopic study ...} \\\colorbox{green!20}{is typically the next} \\\colorbox{green!20}{step in management.}}
  \edge node[midway, right, fill=beaublue] {\textbf{(b)} context}; {\colorbox{green!20}{\textbf{(a)}}\\\colorbox{green!20}{If volvulus is suspected, caution with} \\ \colorbox{green!20}{ non water soluble contrast is mandatory}  \\ \colorbox{green!20}{as ... operative complications.}}
  ]
\end{tikzpicture}
\end{center}

\caption{Semantic hierarchy after the first transformation pass. \colorbox{green!20}{\textbf{(Subtask a)}} The source sentence is split up and rephrased into a set of syntactically simplified sentences. \colorbox{beaublue}{\textbf{(Subtask b)}} Then, the split sentences are connected with information about their constituency type to establish a contextual hierarchy between them. \textbf{\colorbox{red!20}{(Subtask c)}} Finally, by identifying and classifying the rhetorical relation that holds between the simplified sentences, their semantic relationship is preserved. } 
\label{fig:subordination_post_example}
\end{figure}
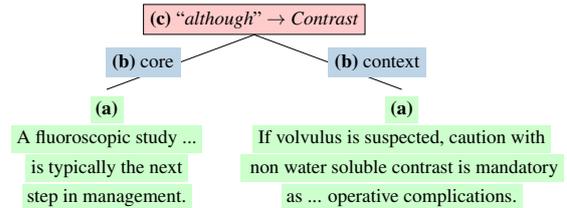

\paragraph{\colorbox{green!20}{(a) Sentence Splitting}} In a first step, the current proposition is decomposed into a set of shorter utterances that present a more regular structure (15, 17). This is achieved through disembedding clausal or phrasal components 
and converting them into stand-alone sentences. Accordingly, the transformation rule encodes both the split point and the rephrasing procedure for reconstructing grammatically sound sentences. Table \ref{rulesAndFrequency} provides an overview of the linguistic constructs that are tackled by our approach.\footnote{Note that this subtask is presented in detail in \citet{niklaus-etal-2019-transforming} and \citet{niklaus-etal-2019-dissim}. 
Therefore, we focus on subtasks b and c in this work.} 
Each split will create two 
sentences with a simpler syntax. They are represented as leaf nodes in the linked proposition tree (18, 19) (see subtask a in Fig. \ref{fig:subordination_post_example}). To establish a semantic hierarchy between the split spans, two further subtasks are carried out. 

\begin{table}[!htb]
\scriptsize
\centering
  \begin{tabular}{ c | l | c  | c  }
    \toprule
    & \textsc{Clausal/Phrasal type} & \textsc{Hierarchy} & \textsc{\# rules}  \\ \hline \hline
    \multicolumn{4}{c} {\textbf{Clausal disembedding}} \\ \hline
    1 & Coordinate clauses & coordinate & 1 \\ \hline

    2 & Adverbial clauses & subordinate & 6  \\ \hline
    
    
    
    3a & Relative clauses (non-restrictive) & subordinate & 5  \\ \hline
    
    3b & Relative clauses (restrictive) & subordinate & 4  \\ \hline
    
    4 & Reported speech & subordinate & 4  \\ \hline\hline
    
    \multicolumn{4}{c} {\textbf{Phrasal disembedding}} \\ \hline 
    5 & Coordinate verb phrases & coordinate & 1 \\ \hline
    6 &  Coordinate noun phrases & coordinate & 2  \\ \hline
     6 &  Participial phrases & subordinate & 4  \\ \hline
    8a &  Appositions (non-restrictive) & subordinate & 1  \\ \hline
    8b & Appositions (restrictive) & subordinate & 1  \\ \hline
    
    9 & Prepositional phrases & subordinate & 3  \\ \hline
    10 & Adjectival and adverbial phrases & subordinate & 2 \\ \hline
    11 & Lead NPs & subordinate & 1  \\ \hline \hline
    
    
    & Total & & 35 \\ \bottomrule

  \end{tabular} 
  
  \caption{Linguistic constructs addressed by our proposed context-preserving TS approach \textsc{DisSim}.}
  \label{rulesAndFrequency}
\end{table}

\paragraph{\colorbox{beaublue}{(b) Constituency Type Classification}} To set up a contextual hierarchy between the split sentences, the transformation rule determines the constituency type $c \in CL$ of the leaf nodes that were created in the previous step (22-24). To differentiate between \textit{core} sentences that contain the key message of the input and \textit{contextual} sentences that provide additional information about it, the transformation pattern encodes a simple syntax-based method. Based on the assumption that subordinations commonly express background information, simplified propositions resulting from subordinate clausal or phrasal elements are classified as context sentences, while those emerging from their superordinate counterparts are labelled as core sentences. Coordinations, too, are flagged as core sentences, since they are of equal status and typically depict the main information of the input (see subtask b in Fig. \ref{fig:subordination_post_example}).\footnote{This approach relates to the concept of nuclearity in RST. For details, see Section \ref{app:constituency} in the appendix.}

\paragraph{\colorbox{red!20}{(c) Rhetorical Relation Identification}} To preserve the semantic relationship between the simplified propositions, we classify the rhetorical relation $rel \in REL$ that holds between them. For this purpose, we utilize a predefined list of rhetorical cue words 
adapted from the work of \citet{Taboada13}.\footnote{The full list of cue phrases that serve as lexical features for the identification of rhetorical relations 
is provided in the appendix in Section \ref{appendix:mapping_cue_phrases}.}
To infer the type of rhetorical relation, the transformation pattern first extracts the cue phrase of the given sentence (26). It is then used as a lexical feature for classifying the type of rhetorical relation that connects the split propositions (27, 28). For example, the rule in Table \ref{examplePatterns} specifies that the phrase \textit{``although''} is the cue word in the source sentence of Fig. \ref{intro_example1}, which is mapped to a ``Contrast'' relationship according to the findings in \newcite{Taboada13} (see subtask c in Figure \ref{fig:subordination_post_example}).

\paragraph{Recursion} Next, the linked proposition tree is updated by replacing the leaf node that was processed in this run with the newly generated subtree (30). It is composed of the simplified propositions, their semantic relationship $rel \in REL$ and constituency labels $c \in CL$. Fig. \ref{fig:subordination_post_example} depicts the result of the first transformation pass on the example sentence from Fig. \ref{intro_example1}. The resulting leaf nodes are then recursively simplified in a top-down fashion (32).


\paragraph{Termination} The algorithm terminates when no more rule matches the set of simplified propositions $prop \in PROP$ in the leaf nodes. It outputs the source sentence's linked proposition tree $LPT$ (36), representing its semantic hierarchy of minimal semantic units. In that way, the input is transformed into a set of hierarchically ordered and semantically interconnected sentences that present a simplified syntax. Figure \ref{intro_example1} shows the final linked proposition tree of our example sentence.

\section{Evaluation}


\subsection{Experimental Setup\footnote{To enable reproducible research, all code / datasets used in the experiments are provided online: \url{https://github.com/Lambda-3/DiscourseSimplification/tree/master/supplemental_material}.}}

\subsubsection{Automatic Metrics}
\label{sec:experimental_setup_subtask2_automatic}

We 
evaluate the constituency type classification and rhetorical relation identification steps by mapping the simplified sentences that were generated in the sentence splitting subtask to the Elementary Discourse Units (EDUs) of the RST-DT corpus.\footnote{\url{https://t1p.de/n6t9}} This dataset is 
a collection of 385 Wall Street Journal articles 
annotated with rhetorical relations based on the RST framework \cite{mann1988rhetorical}. 
For matching simplified sentences generated by our TS approach to the annotations of the RST-DT corpus, we compare each split sentence to all the EDUs of the corresponding input sentence. For each pair, we search for the longest contiguous matching subsequence. Next, based on the size of the matched sequences, a similarity score between the two input strings is calculated. 
Each pair whose similarity score surpasses an empirically determined threshold of 0.65 is considered a match.

\paragraph{Constituency Type Classification} 

To determine whether the hierarchical relationship that is assigned by our TS framework between a pair of simplified sentences is correct, we check if the hierarchy of its contextual layers corresponds to the nuclearity of the aligned text fragments of the RST-DT. For this purpose, we make use of the nuclearity status encoded in the annotations of this dataset. In addition, we compare the performance of our TS approach with that of a set of widely used sentence-level discourse parsers on this task.

\paragraph{Rhetorical Relation Identification}

To assess the performance of the rhetorical relation identification step, we determine the distribution of the relation types allocated by our TS approach when operating on the 7,284 input sentences of the RST-DT and compare it to the distribution of the manually annotated rhetorical relations of this corpus. 
Moreover, we examine for each matching sentence pair whether the rhetorical relation assigned by our TS framework equates the relation that connects the corresponding EDUs in the RST-DT dataset. For this purpose, we apply the more coarse-grained classification scheme from \citet{Taboada13}, who group the full set of 78 rhetorical relations that are used in the RST-DT corpus into 19 classes of relations that share rhetorical meaning. Finally, we analyze the performance of our framework on the relation labeling task in comparison to a number of discourse parser baselines.

\subsubsection{Manual Analysis}
To get a deeper insight into the accuracy of the semantic hierarchy established between the split components, the automatic evaluation described above is complemented by a manual analysis. Three human judges independently of each other assessed each decomposed sentence according to the following four criteria: 
(i) \textbf{Limitation to core information}: Is the simplified output limited to core information of the input sentence? \textit{(yes - no - malformed)}; (ii) \textbf{Soundness of the contextual proposition}: Does the simplified sentence express a meaningful context fact?  \textit{(yes - no)}; (iii) \textbf{Correctness of the context allocation}: Is the contextual sentence assigned to the parent sentence to which it refers? \textit{(yes - no)}; and (iv) \textbf{Properness of the identified semantic relationship}: Is the contextual sentence linked to its parent sentence via the correct semantic relation? \textit{(yes - no - unspecified)}.

The first three categories of our analysis address the correctness of the constituency type classification task, while the latter targets the rhetorical relation identification step.
The annotation task was carried out on 
a random sample of 100 sentences from the OIE2016 Open IE benchmark \cite{stanovsky2016benchmark}.

\subsection{Results}


\subsubsection{Automatic Metrics}

Using the matching function described in Section \ref{sec:experimental_setup_subtask2_automatic}, we obtained 1,827 matched sentence pairs, i.e. 11.74\% of the pairs of simplified sentences were successfully mapped to a counterpart of EDUs from the RST-DT. The relatively low number of matches can be attributed to the fact that the text spans we compare have very different features.\footnote{For details, see Section \ref{app:evaluation} in the appendix.}  
As we are primarily interested in determining whether the constituency and relation labels that are assigned by our approach are correct, we will focus on precision in the following.\footnote{
The fraction of labels that are successfully retrieved (i.e. recall) is of minor importance in our setting. In addition, this score might be biased, since a large proportion of EDUs from RST-DT is not mapped to a counterpart of simplified propositions in our experiments. Therefore, we refrain from reporting recall scores.} 

\paragraph{Constituency Type Classification}

In 88.89\% of the matched sentence pairs, the hierarchical relationship that is allocated between a pair of simplified sentences by our reference TS implementation \textsc{DisSim} corresponds to the nuclearity status of the aligned EDUs from RST-DT, i.e. in case of a nucleus-nucleus relationship in RST-DT, both output sentences from \textsc{DisSim} are assigned to the same context layer, while in case of a nucleus-satellite relationship the sentence mapped to the nucleus EDU is allocated to the context layer \textit{cl}, whereas the sentence mapped to the satellite span is assigned to the subordinate context layer \textit{cl}+1. The majority of the cases where our TS approach assigns a hierarchical relationship that differs from the nuclearity in the RST-DT corpus can be attributed to relative clauses (see Table \ref{tab:identification_examples}). 

\begin{table}[!htb]
\centering
\scriptsize
  \begin{tabular}{ p{1cm} | p{2.6cm} p{2.8cm}  }
    \toprule
    
    Source & \multicolumn{2}{p{5.7cm}}{Mr. Volk, 55 years old, succeeds Duncan Dwight, who retired in September.} \\ \hline
    & \textsc{Mapped Span 1} & \textsc{Mapped Span 2} \\ \hline
    \textsc{DisSim} & Volk succeeds Duncan Dwight. & Duncan Dwight retired in September.\\ 
    \rowcolor{beaublue}Nuclearity & context layer 0 & context layer 1 \\
    \cellcolor{red!20}Relation & \multicolumn{2}{c}{\cellcolor{red!20}{Elaboration}} \\ \hline
    RST-DT & Mr. Volk, 55 years old, succeeds Duncan Dwight, & who retired in September.\\ 
    \rowcolor{beaublue}Nuclearity & nucleus & nucleus \\
    \cellcolor{red!20}Relation & \multicolumn{2}{c}{\cellcolor{red!20}{Elaboration}} \\ \bottomrule 

  \end{tabular} 
  
  \caption{
  Example.}
  \label{tab:identification_examples}
\end{table}

\begin{table}[!htb]
\centering
\footnotesize
  \begin{tabular}{  p{4.25cm} | c c }
    \toprule
     & \cellcolor{beaublue!50}nuclearity & \cellcolor{red!20}relation \\ \hline
    DPLP \cite{ji-eisenstein-2014-representation} & \cellcolor{beaublue!50}71.1 & \cellcolor{red!20}61.8 \\
    \citet{feng-hirst-2014-linear} & \cellcolor{beaublue!50}71.0 & \cellcolor{red!20}58.2 \\
    2-Stage Parser \cite{wang-etal-2017-two} & \cellcolor{beaublue!50}72.4 & \cellcolor{red!20}59.7 \\
    \citet{lin-etal-2019-unified} & \cellcolor{beaublue!50}\textbf{91.3} & \cellcolor{red!20}\textbf{81.7} \\ \hline
    SPADE \cite{soricut-marcu-2003-sentence} & \cellcolor{beaublue!50}56.1 & \cellcolor{red!20}44.9 \\
    HILDA \cite{hernault2010hilda} & \cellcolor{beaublue!50}59.7 & \cellcolor{red!20}48.2 \\
    \textsc{PAR-s} \cite{joty-etal-2015-codra} & \cellcolor{beaublue!50}75.2 & \cellcolor{red!20}66.1 \\
    \citet{lin-etal-2019-unified} & \cellcolor{beaublue!50}(86.4)* & \cellcolor{red!20}(77.5)* \\
    \rowcolor{gray!15}\textsc{DisSim} & \cellcolor{beaublue}\textbf{88.9} & \cellcolor{red!40}\textbf{69.5} \\
    \bottomrule
  \end{tabular} 
  
  \caption{Precision of \textsc{DisSim} and the discourse parser baselines, as reported by their authors. (*) In case of automatic discourse segmentation, for \citet{lin-etal-2019-unified} the F\textsubscript{1}-score is available only.}
  \label{tab:comparison_with_discourse_parsers}
\end{table}

Table \ref{tab:comparison_with_discourse_parsers} displays the precision that the discourse parser baselines achieve on the 991 sentences of the RST-DT test set in distinguishing between nucleus and satellite spans \textit{(``nuclearity'')}. For the approaches in the upper part of the table, the authors report the systems' performance when using gold EDU segmentation, while for those in the lower part the performance is indicated based on automatic segmentation, i.e. when they are fed the output of their respective discourse segmenter. Since our framework makes use of the simplified sentences that were generated in the previous step when setting up the semantic hierarchy, it is better comparable to the latter group. The figures show that in this case our approach outperforms all other systems in the constituency type classification task by a large margin of 13.7\% at a minimum.\footnote{A very recent approach to intra-sentential sentence parsing was proposed in \citet{lin-etal-2019-unified}, achieving an F\textsubscript{1}-score of 86.4\%. However, the authors do not report its precision.}

\paragraph{Rhetorical Relation Identification}
\label{sec:rhetorical_relation_identification_results}

Table \ref{tab:RST_counts_RST-DT} displays the frequency distribution of the 19 classes of rhetorical relations that were specified in \citet{Taboada13}. 
The ten most frequently occurring classes make up for 89.45\% of the relations that are present in the dataset. We decided to limit ourselves to these classes in the evaluation of the rhetorical relation identification step, with two exceptions. First, we did not take into account the ``Topic-change'' and ``Same-unit'' classes. Second, we merged the two highly related classes of ``Cause'' and ``Explanation'' into a single category.

\begin{table}[!htb]
\centering
\footnotesize
  \begin{tabular}{  p{1.8cm}  c  c  c  }
    \toprule
    \textsc{Rhet. Relation} & \textsc{Count} & \textsc{Percent.} & \cellcolor{red!20}\textsc{Precision}\\ \hline\hline
    \textbf{Elaboration} & 7,675 & 25.65\% & \cellcolor{red!20}0.5550 \\ \hline
    \textbf{Joint} & 7,116 & 23.78\% & \cellcolor{red!20}0.6673 \\ \hline
    \textbf{Attribution} & 2,984 & 9.97\% & \cellcolor{red!20}0.9601 \\ \hline
    Same-unit & 2,788 & 9.32\%  & \cellcolor{red!20}---\\ \hline
    \textbf{Contrast} & 1,522 & 5.09\% & \cellcolor{red!20}0.7421 \\ \hline 
    Topic-change & 1,315 & 4.39\% & \cellcolor{red!20}--- \\ \hline
    \textbf{Explanation} & 966 & 3.21\% & \cellcolor{red!20}{} \\ \cline{1-3}
    \textbf{Cause} & 754 & 2.52\% & \multirow{-2}{*}{\cellcolor{red!20}{0.7037}}\\ \hline
    \textbf{Temporal} & 964 & 3.22\% & \cellcolor{red!20}0.7895 \\ \hline
    \textbf{Background} & 897 & 2.30\% & \cellcolor{red!20}0.4459 \\  \hline\hline
    
     &   &  & \cellcolor{red!40}avg.: 0.6948\\ \bottomrule 
     
     \multicolumn{4}{p{0.96\linewidth}}{{\scriptsize{Evaluation (2.0\%), Enablement (1.8\%), Comparison (1.5\%), Textual organization (1.2\%), Condition (1.1\%), Topic-comment (0.9\%), Manner-means (0.7\%), Summary (0.7\%), Span (0.0\%)}}} \\
     
  \end{tabular} 
  
  \caption{Frequency distribution of the 19 classes of 
  relations from \citet{Taboada13} 
  and the precision of \textsc{DisSim}'s rhetorical relation identification step.}
  \label{tab:RST_counts_RST-DT}
\end{table}



The right column in Table \ref{tab:RST_counts_RST-DT} displays the precision of our TS approach for each class of rhetorical relation when run over the sentences from RST-DT. 
The ``Attribution'' relation reaches by far the highest precision. The remaining relations, too, show decent scores, with a precision of around 70\%. The only exception is ``Background''. 
The difficulty with this type of relationship is that it signifies a very broad category that is not signalled by discourse markers and therefore hard to detect by our approach \cite{Taboada13}. 
With an average precision of 69.5\% in the relation labeling task (see Table \ref{tab:comparison_with_discourse_parsers}
), our framework again surpasses all the discourse parser baselines under consideration when using automatic discourse segmentation.\footnote{with the exception of \citet{lin-etal-2019-unified}'s parser, for which only the F\textsubscript{1}-score is reported by the authors, though. Hence, it is not directly comparable to the other approaches whose performance is analyzed based on their precision.}

When comparing the distribution of the rhetorical relations that were identified by our TS approach on the source sentences from the RST-DT (see Figure \ref{fig:distribution_rhetorical_relations_dissim} in Section \ref{app:evaluation} in the appendix) to that of the manually annotated gold relations displayed in Table \ref{tab:RST_counts_RST-DT}, it turns out that there is a very high similarity between the two of them. 
However, it must be noted that in about a fifth of the cases, our TS approach is not able to identify a rhetorical relation between a pair of decomposed sentences \textit{(``Unknown'')}. For the most part, this can be attributed to sentence pairs whose relation is not explicitly stated in the underlying source sentence. Since our approach is based on cue phrases, searching for discourse markers that explicitly signal rhetorical relations, it has difficulties in identifying relations that can merely be implied. 


\subsubsection{Manual Analysis}

The results of the human evaluation are displayed in Table \ref{tab:subtask2_manual_results}. The inter-annotator agreement was calculated using Fleiss' $\kappa$ \cite{fleiss1971mns}. 
The figures indicate fair to substantial agreement between the three annotators, suggesting that the evaluation scores 
present a reliable result. 

\begin{table}[!htb]
\centering
\scriptsize
\begin{tabular}{ p{2.1cm} | c  c  c  c | c}
\toprule
Category & Yes & No & Malf. & Unspec. & $\kappa$ \\
\hline
\rowcolor{beaublue!50}Limitation to core information           & \cellcolor{beaublue}\textbf{68.2\%} & 20.0\% & 11.9\% & --- & \cellcolor{beaublue!20}0.39\\
\rowcolor{beaublue!50}Soundness of the contextual proposition   & \cellcolor{beaublue}\textbf{83.1\%} & 16.9\% & --- & --- & \cellcolor{beaublue!20}0.51 \\
\rowcolor{beaublue!50}Correctness of the context allocation         & \cellcolor{beaublue}\textbf{93.2\%} & 6.8\% & --- & --- & \cellcolor{beaublue!20}0.41 \\
\rowcolor{red!20}Properness of the semantic relationship      & \cellcolor{red!40}\textbf{69.8\%} & 7.0\% & --- & 23.2\% & \cellcolor{red!10}0.69 \\

\bottomrule
\end{tabular}
\caption{Results of the manual analysis.} 
\label{tab:subtask2_manual_results}
\end{table}

In more than two out of three cases, the annotators marked the propositions that were classified as core sentences by our TS approach as correct, thus approving that they have a meaningful interpretation and that their content is truly restricted to core information of the underlying source sentence. Only about 12\% of the simplified sentences are malformed according to our annotations. The remaining fifth of output core sentences was judged as being misclassified, i.e. they rather contribute less relevant background data than key information of the input. 
Regarding the soundness of the context propositions, 
only about 17\% of the output propositions that were classified as context sentences were labelled as being inaccurate, while as many as 83\% present proper contextual propositions, expressing a meaningful context fact that is asserted by the input and can be properly interpreted. 
Furthermore, 93\% of the context sentences are assigned to their respective parent sentence, whereas only 6\% of them are misallocated, according to the annotators' labels. 
Finally, our evaluation revealed that our TS approach shows a decent performance for the rhetorical relation identification step, too. More than two-thirds of the sentence pairs are classified with the correct rhetorical relation, according to our manual analysis. Only 7\% of them are assigned an improper relation. However, in nearly a quarter of the cases, our TS approach was not able to identify a semantic relationship between the given pair of sentences. 
This can be explained by the fact that for this subtask, our framework follows a rather simplistic approach that is primarily based on cue phrases. Therefore, it fails to identify a semantic relationship whenever none of the specified keywords 
appears in the underlying input sentence. As a result, our approach provides very precise results. Covering only a small subset of rhetorical relations 
it lacks in completeness, though.

\subsection{Extrinsic Evaluation}

The fine-grained representation of complex sentences in the form of hierarchically ordered and semantically interconnected propositions may serve as an intermediate representation for downstream tasks.
An application area that may benefit greatly from our approach 
as a preprocessing step is the task of Open IE \cite{Banko07}. 

\paragraph{Coverage and Accuracy}
We demonstrated that relational tuples in the form of predicate-argument structures can be extracted more easily from the simplified sentences \cite{cetto-etal-2018-graphene,cetto-etal-2018-graphene-context}. In fact, a comparative analysis with state-of-the-art Open IE systems \cite{fader-etal-2011-identifying,mausam-etal-2012-open,delcorro-2013-clausie,angeli-etal-2015-leveraging,StanovskyFDG16,Mausam16,gashteovski-etal-2017-minie,saha-etal-2017-bootstrapping,stanovsky2018supervised} on a large benchmark corpus \cite{stanovsky2016benchmark} revealed that their performance in terms of precision and recall is improved by up to 25\% and 27\%, respectively, when taking advantage of the split propositions instead of dealing with the complex source sentences (see Table \ref{tab:improvements_through_preprocessed}).



\begin{table}[!htb]
\centering
\footnotesize
\begin{tabular}{ p{1.3cm} | c  c  c  c }
\toprule
System & Precision & Recall & F\textsubscript{1} & AUC \\
\hline
\textsc{ReVerb}            & \cellcolor{amber(sae/ece)!50}{-7.2\%} & \cellcolor{ao(english)!40}+19.1\% & \cellcolor{ao(english)!40}+8.4\% & \cellcolor{ao(english)!40}\textbf{+39.2}\% \\
\textsc{Ollie}             & \cellcolor{ao(english)!40}+1.1\% & \cellcolor{amber(sae/ece)!50}-1.5\% & \cellcolor{amber(sae/ece)!50}-0.3\% & \cellcolor{amber(sae/ece)!50}-1.1\% \\
ClausIE               & \cellcolor{ao(english)!40}+17.0\% & \cellcolor{amber(sae/ece)!50}-3.5\% & \cellcolor{ao(english)!40}+8.1\% &  \cellcolor{ao(english)!40}+13.0\%        \\
Stanford Open IE      & \cellcolor{ao(english)!40}\textbf{+25.0\%} & \cellcolor{ao(english)!40}\textbf{+27.2\%} & \cellcolor{ao(english)!40}\textbf{+25.5\%} & \cellcolor{ao(english)!40}+35.5\%\\
PropS                      & \cellcolor{amber(sae/ece)!50}-6.1\% & \cellcolor{ao(english)!40}+16.9\% & \cellcolor{ao(english)!40}+4.5\% & \cellcolor{ao(english)!40}+12.4\%\\
OpenIE-4              & \cellcolor{ao(english)!40}+10.0\% & \cellcolor{ao(english)!40}+8.6\% & \cellcolor{ao(english)!40}+9.4\% & \cellcolor{ao(english)!40}+19.6\% \\
MinIE                & \cellcolor{ao(english)!40}+23.7\% & \cellcolor{amber(sae/ece)!50}-1.8\% & \cellcolor{ao(english)!40}+12.5\% & \cellcolor{ao(english)!40}+21.3\%\\
OpenIE-5              & \cellcolor{ao(english)!40}+5.0\% & \cellcolor{ao(english)!40}+4.2\% & \cellcolor{ao(english)!40}+4.6\% & \cellcolor{ao(english)!40}+9.0\%\\
RnnOIE               & \cellcolor{amber(sae/ece)!50}-15.0\% & \cellcolor{ao(english)!40}+0.9\% & \cellcolor{amber(sae/ece)!50}-8.3\% & \cellcolor{amber(sae/ece)!50}-14.1\%   \\

\bottomrule
\end{tabular}
\caption{Improvements in precision, recall, F\textsubscript{1} and AUC on the OIE2016 benchmark when using our reference 
implementation \textsc{DisSim} as a pre-processing step.}
\label{tab:improvements_through_preprocessed}
\end{table}

\paragraph{Coherence}
Moreover, our TS approach enables existing Open IE systems to enrich their output with semantic information. 
By capturing intra-sentential rhetorical structures and hierarchical relationships between the extracted tuples, the shallow semantic representation of Open IE tuples in the form of isolated predicate-argument structures is extended. In that way, the semantic context of the individual tuples is preserved.\footnote{For details, see Section \ref{app:merits} in the appendix.}


\section{Related Work}






\paragraph{Discourse-level TS}
The vast majority of structural TS approaches do not take into account discourse-level aspects. 
However, two notable exceptions have to be mentioned.
\newcite{siddharthan2006syntactic} was the first to use discourse-aware cues in the simplification process. 
As opposed to our approach, though, where a semantic relationship is established for each simplified output sentence, only a comparatively low number of sentences is linked by such cue words. 
Another approach that operates on the level of discourse 
was proposed by \newcite{stajner2017leveraging}. It performs a semantically motivated content reduction 
by maintaining only those parts of a sentence that belong to factual event mentions.
Our approach, on the contrary, aims to preserve all the information contained in the source. 

\paragraph{Discourse Parsing}

The challenge of uncovering coherence structures in texts is pursued in the field of Discourse Parsing. It aims to identify discourse relations that hold between textual units in a document \cite{marcu-1997-rhetorical}. 
A well-established theory of text structure used in this area is RST. Here,
textual coherence is explained by the existence of rhetorical relations that hold between 
adjacent text spans in a hierarchical structure. 
Approaches to 
detect rhetorical structure arrangements in texts range from early rule-based approaches \cite{marcu-2000-rhetorical} to supervised data-driven models that were trained on annotated corpora such as the RST-DT \cite{feng-hirst-2014-linear,li-etal-2014-recursive,lin-etal-2019-unified}.\footnote{Section \ref{app:related_work} in the appendix elaborates on why it is not possible to simply use an RST parser for establishing the semantic hierarchy between the decomposed spans.}

\section{Conclusion}
We presented a context-preserving TS approach that transforms structurally complex sentences into a hierarchical representation in the form of core sentences and accompanying contexts that are semantically linked by rhetorical relations.
In our experiments, we mapped the simplified sentences from our reference implementation \textsc{DisSim} to the EDUs from 
RST-DT 
and 
 showed that we obtain a very high precision of 89\% for the constituency type classification and a decent score of 69\% on average for the rhetorical relation identification. 
In the future, we plan to improve the latter step by extending our approach to also capture implicit relationships between the decomposed sentences.

\bibliographystyle{acl_natbib}
\bibliography{anthology,acl2021}

\begin{thebibliography}{49}
\expandafter\ifx\csname natexlab\endcsname\relax\def\natexlab#1{#1}\fi

\bibitem[{Abend and Rappoport(2013)}]{abend-rappoport-2013-universal}
Omri Abend and Ari Rappoport. 2013.
\newblock \href {https://www.aclweb.org/anthology/P13-1023} {{U}niversal
  {C}onceptual {C}ognitive {A}nnotation ({UCCA})}.
\newblock In \emph{Proceedings of the 51st Annual Meeting of the Association
  for Computational Linguistics (Volume 1: Long Papers)}, pages 228--238,
  Sofia, Bulgaria. Association for Computational Linguistics.

\bibitem[{Angeli et~al.(2015)Angeli, Johnson~Premkumar, and
  Manning}]{angeli-etal-2015-leveraging}
Gabor Angeli, Melvin~Jose Johnson~Premkumar, and Christopher~D. Manning. 2015.
\newblock \href {https://doi.org/10.3115/v1/P15-1034} {Leveraging linguistic
  structure for open domain information extraction}.
\newblock In \emph{Proceedings of the 53rd Annual Meeting of the Association
  for Computational Linguistics and the 7th International Joint Conference on
  Natural Language Processing (Volume 1: Long Papers)}, pages 344--354,
  Beijing, China. Association for Computational Linguistics.

\bibitem[{Bach et~al.(2011)Bach, Gao, Vogel, and Waibel}]{bach-etal-2011-tris}
Nguyen Bach, Qin Gao, Stephan Vogel, and Alex Waibel. 2011.
\newblock \href {https://www.aclweb.org/anthology/I11-1053} {{T}ri{S}: A
  statistical sentence simplifier with log-linear models and margin-based
  discriminative training}.
\newblock In \emph{Proceedings of 5th International Joint Conference on Natural
  Language Processing}, pages 474--482, Chiang Mai, Thailand. Asian Federation
  of Natural Language Processing.

\bibitem[{Banko et~al.(2007)Banko, Cafarella, Soderland, Broadhead, and
  Etzioni}]{Banko07}
Michele Banko, Michael~J. Cafarella, Stephen Soderland, Matt Broadhead, and
  Oren Etzioni. 2007.
\newblock Open information extraction from the web.
\newblock In \emph{Proceedings of the 20th International Joint Conference on
  Artifical Intelligence}, pages 2670--2676, San Francisco, CA, USA. Morgan
  Kaufmann Publishers Inc.

\bibitem[{Carlson and Marcu(2001)}]{carlson2001discourse}
Lynn Carlson and Daniel Marcu. 2001.
\newblock \href {https://www.isi.edu/~marcu/discourse/tagging-ref-manual.pdf}
  {Discourse tagging reference manual}.
\newblock \emph{ISI Technical Report ISI-TR-545}, 54:56.

\bibitem[{Cetto et~al.(2018{\natexlab{a}})Cetto, Niklaus, Freitas, and
  Handschuh}]{cetto-etal-2018-graphene-context}
Matthias Cetto, Christina Niklaus, Andr{\'e} Freitas, and Siegfried Handschuh.
  2018{\natexlab{a}}.
\newblock \href {https://www.aclweb.org/anthology/C18-2021} {{G}raphene: a
  context-preserving open information extraction system}.
\newblock In \emph{Proceedings of the 27th International Conference on
  Computational Linguistics: System Demonstrations}, pages 94--98, Santa Fe,
  New Mexico. Association for Computational Linguistics.

\bibitem[{Cetto et~al.(2018{\natexlab{b}})Cetto, Niklaus, Freitas, and
  Handschuh}]{cetto-etal-2018-graphene}
Matthias Cetto, Christina Niklaus, Andr{\'e} Freitas, and Siegfried Handschuh.
  2018{\natexlab{b}}.
\newblock \href {https://www.aclweb.org/anthology/C18-1195} {{G}raphene:
  Semantically-linked propositions in open information extraction}.
\newblock In \emph{Proceedings of the 27th International Conference on
  Computational Linguistics}, pages 2300--2311, Santa Fe, New Mexico, USA.
  Association for Computational Linguistics.

\bibitem[{Del~Corro and Gemulla(2013)}]{delcorro-2013-clausie}
Luciano Del~Corro and Rainer Gemulla. 2013.
\newblock \href {https://doi.org/10.1145/2488388.2488420} {Clausie:
  Clause-based open information extraction}.
\newblock In \emph{Proceedings of the 22nd International Conference on World
  Wide Web}, WWW ’13, page 355–366, New York, NY, USA. Association for
  Computing Machinery.

\bibitem[{Evans and Or\v{a}san(2019)}]{evansorasan2019}
Richard Evans and Constantin Or\v{a}san. 2019.
\newblock \href {https://doi.org/10.1017/S1351324918000384} {Identifying signs
  of syntactic complexity for rule-based sentence simplification}.
\newblock \emph{Natural Language Engineering}, 25(1):69–119.

\bibitem[{Fader et~al.(2011)Fader, Soderland, and
  Etzioni}]{fader-etal-2011-identifying}
Anthony Fader, Stephen Soderland, and Oren Etzioni. 2011.
\newblock \href {https://www.aclweb.org/anthology/D11-1142} {Identifying
  relations for open information extraction}.
\newblock In \emph{Proceedings of the 2011 Conference on Empirical Methods in
  Natural Language Processing}, pages 1535--1545, Edinburgh, Scotland, UK.
  Association for Computational Linguistics.

\bibitem[{Fay(1990)}]{collinsgrammar}
Richard Fay, editor. 1990.
\newblock \emph{Collins Cobuild English Grammar}.
\newblock Collins.

\bibitem[{Feng and Hirst(2014)}]{feng-hirst-2014-linear}
Vanessa~Wei Feng and Graeme Hirst. 2014.
\newblock \href {https://doi.org/10.3115/v1/P14-1048} {A linear-time bottom-up
  discourse parser with constraints and post-editing}.
\newblock In \emph{Proceedings of the 52nd Annual Meeting of the Association
  for Computational Linguistics (Volume 1: Long Papers)}, pages 511--521,
  Baltimore, Maryland. Association for Computational Linguistics.

\bibitem[{Ferr{\'{e}}s et~al.(2016)Ferr{\'{e}}s, Marimon, Saggion, and
  AbuRa'ed}]{ferres-2016-yats}
Daniel Ferr{\'{e}}s, Montserrat Marimon, Horacio Saggion, and Ahmed AbuRa'ed.
  2016.
\newblock \href {https://doi.org/10.1007/978-3-319-41754-7\_32} {{YATS:} yet
  another text simplifier}.
\newblock In \emph{Natural Language Processing and Information Systems - 21st
  International Conference on Applications of Natural Language to Information
  Systems, {NLDB} 2016, Salford, UK, June 22-24, 2016, Proceedings}, volume
  9612 of \emph{Lecture Notes in Computer Science}, pages 335--342. Springer.

\bibitem[{Fleiss(1971)}]{fleiss1971mns}
J.L. Fleiss. 1971.
\newblock {Measuring nominal scale agreement among many raters}.
\newblock \emph{Psychological Bulletin}, 76(5):378--382.

\bibitem[{Gashteovski et~al.(2017)Gashteovski, Gemulla, and del
  Corro}]{gashteovski-etal-2017-minie}
Kiril Gashteovski, Rainer Gemulla, and Luciano del Corro. 2017.
\newblock \href {https://doi.org/10.18653/v1/D17-1278} {{M}in{IE}: Minimizing
  facts in open information extraction}.
\newblock In \emph{Proceedings of the 2017 Conference on Empirical Methods in
  Natural Language Processing}, pages 2630--2640, Copenhagen, Denmark.
  Association for Computational Linguistics.

\bibitem[{Heilman and Smith(2010)}]{heilman-2010-extracting}
Michael Heilman and Noah~A. Smith. 2010.
\newblock Extracting simplified statements for factual question generation.
\newblock In \emph{Proceedings of the QG2010: The Third Workshop on Question
  Generation}, pages 11--20.

\bibitem[{Hernault et~al.(2010)Hernault, Prendinger, Ishizuka
  et~al.}]{hernault2010hilda}
Hugo Hernault, Helmut Prendinger, Mitsuru Ishizuka, et~al. 2010.
\newblock Hilda: A discourse parser using support vector machine
  classification.
\newblock \emph{Dialogue \& Discourse}, 1(3).

\bibitem[{Ji and Eisenstein(2014)}]{ji-eisenstein-2014-representation}
Yangfeng Ji and Jacob Eisenstein. 2014.
\newblock \href {https://doi.org/10.3115/v1/P14-1002} {Representation learning
  for text-level discourse parsing}.
\newblock In \emph{Proceedings of the 52nd Annual Meeting of the Association
  for Computational Linguistics (Volume 1: Long Papers)}, pages 13--24,
  Baltimore, Maryland. Association for Computational Linguistics.

\bibitem[{Joty et~al.(2015)Joty, Carenini, and Ng}]{joty-etal-2015-codra}
Shafiq Joty, Giuseppe Carenini, and Raymond~T. Ng. 2015.
\newblock \href {https://doi.org/10.1162/COLI_a_00226} {{CODRA}: A novel
  discriminative framework for rhetorical analysis}.
\newblock \emph{Computational Linguistics}, 41(3):385--435.

\bibitem[{Levy and Andrew(2006)}]{levy-andrew-2006-tregex}
Roger Levy and Galen Andrew. 2006.
\newblock \href {http://www.lrec-conf.org/proceedings/lrec2006/pdf/513_pdf.pdf}
  {Tregex and tsurgeon: tools for querying and manipulating tree data
  structures}.
\newblock In \emph{Proceedings of the Fifth International Conference on
  Language Resources and Evaluation ({LREC}{'}06)}, Genoa, Italy. European
  Language Resources Association (ELRA).

\bibitem[{Li et~al.(2014)Li, Li, and Hovy}]{li-etal-2014-recursive}
Jiwei Li, Rumeng Li, and Eduard Hovy. 2014.
\newblock \href {https://doi.org/10.3115/v1/D14-1220} {Recursive deep models
  for discourse parsing}.
\newblock In \emph{Proceedings of the 2014 Conference on Empirical Methods in
  Natural Language Processing ({EMNLP})}, pages 2061--2069, Doha, Qatar.
  Association for Computational Linguistics.

\bibitem[{Lin et~al.(2019)Lin, Joty, Jwalapuram, and
  Bari}]{lin-etal-2019-unified}
Xiang Lin, Shafiq Joty, Prathyusha Jwalapuram, and M~Saiful Bari. 2019.
\newblock \href {https://doi.org/10.18653/v1/P19-1410} {A unified linear-time
  framework for sentence-level discourse parsing}.
\newblock In \emph{Proceedings of the 57th Annual Meeting of the Association
  for Computational Linguistics}, pages 4190--4200, Florence, Italy.
  Association for Computational Linguistics.

\bibitem[{Mallinson and Lapata(2019)}]{mallinson2019controllable}
Jonathan Mallinson and Mirella Lapata. 2019.
\newblock Controllable sentence simplification: Employing syntactic and lexical
  constraints.
\newblock \emph{arXiv preprint arXiv:1910.04387}.

\bibitem[{Mann and Thompson(1988)}]{mann1988rhetorical}
William~C Mann and Sandra~A Thompson. 1988.
\newblock Rhetorical structure theory: Toward a functional theory of text
  organization.
\newblock \emph{Text-Interdisciplinary Journal for the Study of Discourse},
  8(3):243--281.

\bibitem[{Marcu(1997)}]{marcu-1997-rhetorical}
Daniel Marcu. 1997.
\newblock \href {https://doi.org/10.3115/976909.979630} {The rhetorical parsing
  of unrestricted natural language texts}.
\newblock In \emph{35th Annual Meeting of the Association for Computational
  Linguistics and 8th Conference of the {E}uropean Chapter of the Association
  for Computational Linguistics}, pages 96--103, Madrid, Spain. Association for
  Computational Linguistics.

\bibitem[{Marcu(2000)}]{marcu-2000-rhetorical}
Daniel Marcu. 2000.
\newblock \href {https://www.aclweb.org/anthology/J00-3005} {The rhetorical
  parsing of unrestricted texts: a surface-based approach}.
\newblock \emph{Computational Linguistics}, 26(3):395--448.

\bibitem[{{Mausam} et~al.(2012){Mausam}, Schmitz, Soderland, Bart, and
  Etzioni}]{mausam-etal-2012-open}
{Mausam}, Michael Schmitz, Stephen Soderland, Robert Bart, and Oren Etzioni.
  2012.
\newblock \href {https://www.aclweb.org/anthology/D12-1048} {Open language
  learning for information extraction}.
\newblock In \emph{Proceedings of the 2012 Joint Conference on Empirical
  Methods in Natural Language Processing and Computational Natural Language
  Learning}, pages 523--534, Jeju Island, Korea. Association for Computational
  Linguistics.

\bibitem[{Mausam(2016)}]{Mausam16}
Mausam Mausam. 2016.
\newblock Open information extraction systems and downstream applications.
\newblock In \emph{Proceedings of the Twenty-Fifth International Joint
  Conference on Artificial Intelligence}, IJCAI’16, page 4074–4077. AAAI
  Press.

\bibitem[{Mitkov and Saggion(2018)}]{Saggion2018book}
Ruslan Mitkov and Horacio Saggion. 2018.
\newblock \href
  {https://www.oxfordhandbooks.com/view/10.1093/oxfordhb/9780199573691.001.0001/oxfordhb-9780199573691-e-52}
  {Text simplification}.

\bibitem[{Niklaus et~al.(2018)Niklaus, Cetto, Freitas, and
  Handschuh}]{niklaus-etal-2018-survey}
Christina Niklaus, Matthias Cetto, Andr{\'e} Freitas, and Siegfried Handschuh.
  2018.
\newblock \href {https://www.aclweb.org/anthology/C18-1326} {A survey on open
  information extraction}.
\newblock In \emph{Proceedings of the 27th International Conference on
  Computational Linguistics}, pages 3866--3878, Santa Fe, New Mexico, USA.
  Association for Computational Linguistics.

\bibitem[{Niklaus et~al.(2019{\natexlab{a}})Niklaus, Cetto, Freitas, and
  Handschuh}]{niklaus-etal-2019-dissim}
Christina Niklaus, Matthias Cetto, Andr{\'e} Freitas, and Siegfried Handschuh.
  2019{\natexlab{a}}.
\newblock \href {https://doi.org/10.18653/v1/W19-8662} {{D}is{S}im: A
  discourse-aware syntactic text simplification framework for {E}nglish and
  {G}erman}.
\newblock In \emph{Proceedings of the 12th International Conference on Natural
  Language Generation}, pages 504--507, Tokyo, Japan. Association for
  Computational Linguistics.

\bibitem[{Niklaus et~al.(2019{\natexlab{b}})Niklaus, Cetto, Freitas, and
  Handschuh}]{niklaus-etal-2019-transforming}
Christina Niklaus, Matthias Cetto, Andr{\'e} Freitas, and Siegfried Handschuh.
  2019{\natexlab{b}}.
\newblock \href {https://doi.org/10.18653/v1/P19-1333} {Transforming complex
  sentences into a semantic hierarchy}.
\newblock In \emph{Proceedings of the 57th Annual Meeting of the Association
  for Computational Linguistics}, pages 3415--3427, Florence, Italy.
  Association for Computational Linguistics.

\bibitem[{Quirk et~al.(1985)Quirk, Greenbaum, Leech, and Svartvik}]{Quir85}
Randolph Quirk, Sidney Greenbaum, Geoffrey Leech, and Jan Svartvik. 1985.
\newblock \emph{A Comprehensive Grammar of the English Language}.
\newblock Longman, London.

\bibitem[{Saggion et~al.(2015)Saggion, \v{S}tajner, Bott, Mille, Rello, and
  Drndarevic}]{saggion-2015-simplext}
Horacio Saggion, Sanja \v{S}tajner, Stefan Bott, Simon Mille, Luz Rello, and
  Biljana Drndarevic. 2015.
\newblock \href {https://doi.org/10.1145/2738046} {Making it simplext:
  Implementation and evaluation of a text simplification system for spanish}.
\newblock \emph{ACM Transactions on Accessible Computing}, 6(4).

\bibitem[{Saha and {Mausam}(2018)}]{saha-mausam-2018-open}
Swarnadeep Saha and {Mausam}. 2018.
\newblock \href {https://www.aclweb.org/anthology/C18-1194} {Open information
  extraction from conjunctive sentences}.
\newblock In \emph{Proceedings of the 27th International Conference on
  Computational Linguistics}, pages 2288--2299, Santa Fe, New Mexico, USA.
  Association for Computational Linguistics.

\bibitem[{Saha et~al.(2017)Saha, Pal, and
  {Mausam}}]{saha-etal-2017-bootstrapping}
Swarnadeep Saha, Harinder Pal, and {Mausam}. 2017.
\newblock \href {https://doi.org/10.18653/v1/P17-2050} {Bootstrapping for
  numerical open {IE}}.
\newblock In \emph{Proceedings of the 55th Annual Meeting of the Association
  for Computational Linguistics (Volume 2: Short Papers)}, pages 317--323,
  Vancouver, Canada. Association for Computational Linguistics.

\bibitem[{Siddharthan(2002)}]{siddharthan2002architecture}
Advaith Siddharthan. 2002.
\newblock An architecture for a text simplification system.
\newblock In \emph{Language Engineering Conference, 2002. Proceedings}, pages
  64--71. IEEE.

\bibitem[{Siddharthan(2006)}]{siddharthan2006syntactic}
Advaith Siddharthan. 2006.
\newblock Syntactic simplification and text cohesion.
\newblock \emph{Research on Language and Computation}, 4(1):77--109.

\bibitem[{Siddharthan(2014)}]{siddharthan2014survey}
Advaith Siddharthan. 2014.
\newblock A survey of research on text simplification.
\newblock \emph{ITL-International Journal of Applied Linguistics},
  165(2):259--298.

\bibitem[{Siddharthan and Mandya(2014)}]{Siddharthan2014}
Advaith Siddharthan and Angrosh Mandya. 2014.
\newblock \href {https://doi.org/10.3115/v1/E14-1076} {Hybrid text
  simplification using synchronous dependency grammars with hand-written and
  automatically harvested rules}.
\newblock In \emph{Proceedings of the 14th Conference of the {E}uropean Chapter
  of the Association for Computational Linguistics}, pages 722--731,
  Gothenburg, Sweden. Association for Computational Linguistics.

\bibitem[{Soricut and Marcu(2003)}]{soricut-marcu-2003-sentence}
Radu Soricut and Daniel Marcu. 2003.
\newblock \href {https://www.aclweb.org/anthology/N03-1030} {Sentence level
  discourse parsing using syntactic and lexical information}.
\newblock In \emph{Proceedings of the 2003 Human Language Technology Conference
  of the North {A}merican Chapter of the Association for Computational
  Linguistics}, pages 228--235.

\bibitem[{{\v{S}}tajner and Glava{\v{s}}(2017)}]{stajner2017leveraging}
Sanja {\v{S}}tajner and Goran Glava{\v{s}}. 2017.
\newblock Leveraging event-based semantics for automated text simplification.
\newblock \emph{Expert systems with applications}, 82:383--395.

\bibitem[{{\v{S}}tajner and Popovic(2016)}]{stajner-popovic-2016-text}
Sanja {\v{S}}tajner and Maja Popovic. 2016.
\newblock \href {https://www.aclweb.org/anthology/W16-3411} {Can text
  simplification help machine translation?}
\newblock In \emph{Proceedings of the 19th Annual Conference of the {E}uropean
  Association for Machine Translation}, pages 230--242.

\bibitem[{{\v{S}}tajner and Popovi{\'c}(2018)}]{stajner-popovic-2018-improving}
Sanja {\v{S}}tajner and Maja Popovi{\'c}. 2018.
\newblock \href {https://doi.org/10.18653/v1/W18-7006} {Improving machine
  translation of {E}nglish relative clauses with automatic text
  simplification}.
\newblock In \emph{Proceedings of the 1st Workshop on Automatic Text Adaptation
  ({ATA})}, pages 39--48, Tilburg, the Netherlands. Association for
  Computational Linguistics.

\bibitem[{Stanovsky and Dagan(2016)}]{stanovsky2016benchmark}
Gabriel Stanovsky and Ido Dagan. 2016.
\newblock \href {https://doi.org/10.18653/v1/D16-1252} {Creating a large
  benchmark for open information extraction}.
\newblock In \emph{Proceedings of the 2016 Conference on Empirical Methods in
  Natural Language Processing}, pages 2300--2305, Austin, Texas. Association
  for Computational Linguistics.

\bibitem[{Stanovsky et~al.(2016)Stanovsky, Ficler, Dagan, and
  Goldberg}]{StanovskyFDG16}
Gabriel Stanovsky, Jessica Ficler, Ido Dagan, and Yoav Goldberg. 2016.
\newblock \href {http://arxiv.org/abs/1603.01648} {Getting more out of syntax
  with props}.
\newblock \emph{CoRR}, abs/1603.01648.

\bibitem[{Stanovsky et~al.(2018)Stanovsky, Michael, Zettlemoyer, and
  Dagan}]{stanovsky2018supervised}
Gabriel Stanovsky, Julian Michael, Luke Zettlemoyer, and Ido Dagan. 2018.
\newblock \href {https://doi.org/10.18653/v1/N18-1081} {Supervised open
  information extraction}.
\newblock In \emph{Proceedings of the 2018 Conference of the North {A}merican
  Chapter of the Association for Computational Linguistics: Human Language
  Technologies, Volume 1 (Long Papers)}, pages 885--895, New Orleans,
  Louisiana. Association for Computational Linguistics.

\bibitem[{Taboada and Das(2013)}]{Taboada13}
Maite Taboada and Debopam Das. 2013.
\newblock Annotation upon annotation: Adding signalling information to a corpus
  of discourse relations.
\newblock \emph{D\&D}, 4(2):249--281.

\bibitem[{Wang et~al.(2017)Wang, Li, and Wang}]{wang-etal-2017-two}
Yizhong Wang, Sujian Li, and Houfeng Wang. 2017.
\newblock \href {https://doi.org/10.18653/v1/P17-2029} {A two-stage parsing
  method for text-level discourse analysis}.
\newblock In \emph{Proceedings of the 55th Annual Meeting of the Association
  for Computational Linguistics (Volume 2: Short Papers)}, pages 184--188,
  Vancouver, Canada. Association for Computational Linguistics.

\end{thebibliography}

\appendix

\section{Process of Development of the Transformation Patterns}
\label{app:pattern_development}

One of the fundamental goals of our context-preserving TS approach is to decompose complex assertions into a set of self-contained minimal propositions. Intuitively, the simplified propositions are \textit{``easy to read and understand, and arguably easily processed by computers''} \cite{bach-etal-2011-tris}. Below, we present a more formal specification of the minimality property we aim for in the resulting simplified sentences, both on a syntactic and a semantic level.




\paragraph{Minimality on the Syntactic Level}
In syntax, four types of sentence structures are distinguished: \textit{simple sentences, compound sentences, complex sentences and compound-complex sentences} \cite{Quir85}. 
The goal of our approach is to transform a given source sentence into a set of \textit{simple sentences}. A simple sentence is a sentence that comprises exactly one independent clause, i.e. a group of words that has both a subject (S) and a verb (V), and optionally an indirect or direct object (O), an adverbial (A) or a complement (C). It expresses a complete thought, i.e. some coherent piece of information, as the following example shows: \textit{``I admire her reasoning.''}

Accordingly, the method we propose aims to split and rephrase complex multi-clause sentences into sequences of simple sentences that each contain exactly one independent clause. \citet{collinsgrammar} distinguishes four central functional categories of clauses:

\begin{compactitem}
    \item \textbf{coordinate clauses} (1 rule),
    \item \textbf{adverbial clauses} (6 rules addressing pre- and postposed adverbial clauses as well as adverbial clauses of purpose introduced by the phrase \textit{``to do''}),
    \item \textbf{relative clauses} (9 rules targeting both restrictive and non-restrictive relative clauses, lexicalized on the different types of relative pronouns) and
    \item \textbf{reported speech} (4 rules addressing both direct and reported speech with pre- and postposed attribution).
\end{compactitem}

Thus, in order to divide complex assertions into single-clause elements, we specified, in a first step, a set of transformation patterns that target above-mentioned categories of clauses, converting complex and compound sentences into a set of simple sentences.

However, simple sentences may still include optional constituents that render it overly complex. For instance, the two adverbials \textit{``in Princeton''} and \textit{``in 1955''} in the simple sentence \textit{``[Albert Einstein]\textsubscript{S} [died]\textsubscript{V} [in Princeton]\textsubscript{A} [in 1955]\textsubscript{A}.''} specify additional contextual information that can be left out without producing ill-formed output. Rather, the remaining clause \textit{``Albert Einstein died.''} still carries semantically meaningful information. Hence, in order to transform simple sentences into atomic semantic units, we need to further reduce them to their corresponding clause type. According to \citet{Quir85}, clauses can be classified into seven different clause types based on the grammatical function of their constituents, as illustrated in Table \ref{tab:clause_types} \cite{delcorro-2013-clausie}.

\begin{table*}[!htb]
\footnotesize
\centering
  \begin{tabular}{ c | p{1.5cm} | p{12cm} }
    \toprule 
    & \textsc{Clause Type} & \textsc{Example}  \\ \hline
    
    \textit{T\textsubscript{1}} & \textbf{SV} & \textit{[Albert Einstein]\textsubscript{S} [died]\textsubscript{V}.} \\
    \textit{T\textsubscript{2}} & \textbf{SVA} & \textit{[Albert Einstein]\textsubscript{S} [remained]\textsubscript{V} [in Princeton]\textsubscript{A}.}  \\
    \textit{T\textsubscript{3}} & \textbf{SVC} & \textit{[Albert Einstein]\textsubscript{S} [is]\textsubscript{V} [smart]\textsubscript{C}.}  \\
    \textit{T\textsubscript{4}} & \textbf{SVO} & \textit{[Albert Einstein]\textsubscript{S} [has won]\textsubscript{V} [the Nobel Prize]\textsubscript{O}.}  \\
    \textit{T\textsubscript{5}} & \textbf{SVOO} & \textit{[The Royal Swedish Academy of Sciences]\textsubscript{S} [gave]\textsubscript{V} [Albert Einstein]\textsubscript{O} [the Nobel Prize]\textsubscript{O}.}  \\
    \textit{T\textsubscript{6}} & \textbf{SVOA} & \textit{[The doorman]\textsubscript{S} [showed]\textsubscript{V} [Albert Einstein]\textsubscript{O} [to his office]\textsubscript{A}.}  \\
    \textit{T\textsubscript{7}} & \textbf{SVOC} & \textit{[Albert Einstein]\textsubscript{S} [declared]\textsubscript{V} [the meeting]\textsubscript{O} [open]\textsubscript{C}.}  \\
    
     \bottomrule 

  \end{tabular} 
  
  \caption[The seven types of clauses.]{The seven types of clauses. S: Subject, V: Verb, C: Complement, O: Object, A: Adverbial.}
  \label{tab:clause_types}
\end{table*}

The clause type conveys the \textit{minimal unit of coherent information} in the clause. Accordingly, if a constituent of a clause that is also part of its type is removed, the resulting clause does not carry semantically meaningful information any more \cite{delcorro-2013-clausie}. Hence, while constituents that belong to the clause type are essential components of the corresponding simple sentence, all other constituents are optional and can be discarded without leading to an incoherent or semantically meaningless output. Therefore, we do not only perform clausal disembedding. Instead, we go down to the level of individual phrases, with the goal of extracting and converting optional phrasal constituents into stand-alone sentences.

So far, a minimal proposition can be defined as a simple sentence that is reduced
to its clause type by omitting all optional constituents, i.e. all elements that do
not appear in the type of the underlying clause. However, transforming complex
source sentences into simple sentences and trimming them to their clause types may
still result in over-specified propositions due to overly complex subclausal units. 
For instance, in the simple sentence \textit{``[Bell, a telecommunication company,]\textsubscript{S} [makes and distributes]\textsubscript{V} [electrical goods, computers and building products]\textsubscript{O}.''} (of clause type SVO), the phrasal elements in all three positions are unnecessarily complex. While the subject contains an appositive phrase that further specifies the noun to which it refers \textit{(“Bell”)}, both the verb and the object include coordinated conjunctions that can be decomposed into separate elements, resulting in a set of clauses that present a much simpler syntax. Consequently, we further simplify such utterances by extracting phrasal elements from the input and transforming them into stand-alone sentences.
In total, following \citet{heilman-2010-extracting}, our TS approach addresses seven types of phrasal constructs:

\begin{compactitem}
    \item \textbf{coordinate verb phrases} (1 rule),
    \item \textbf{coordinate noun phrase lists} (2 rules targeting lists of noun phrases in subject and object position),
    \item \textbf{participial phrases} (4 rules addressing both restrictive and non-restrictive embedded, pre- and postposed participial phrases),
    \item \textbf{appositive phrases} (2 rules addressing restrictive and non-restrictive appositions),
    \item \textbf{prepositional phrases} (3 rules addressing prepositional phrases that act as complements of verb phrases and prepositional phrases that are offset by commas),
    \item \textbf{adjectival and adverbial phrases} (2 rules targeting pre- and postposed adjectival and adverbial phrases) and
    \item \textbf{lead noun phrases} (1 rule).
\end{compactitem}


To sum up, by the notion of a minimal proposition we understand a simple sentence that has been broken down to its essential constituents, i.e. the elements that are part of its clause type, while also extracting a specified set of phrasal units.

\paragraph{Minimality on the Semantic Level}

From a semantic point of view, a text can be seen as a collection of events (also called ``frames'' or ``scenes'') \cite{abend-rappoport-2013-universal} that describe some activity, state or property. From a semantic perspective, the goal of our TS approach is to split sentences into separate frames, where each of them presents a single event.
An event typically consists of a predicate, a set of arguments and optionally secondary relations. While the predicate (typically a verb, but nominal or adjectival
predicates are also possible) is the main determinant of what the event is about,
arguments describe its participants (e.g., who and where). They represent core elements of a frame, i.e. essential components that make it unique and different from
other frames. Secondary relations, in contrast, represent non-core elements that
introduce additional relations, describing further event properties (e.g., when and
how).
In order to avoid over-specified propositions that are difficult to handle for downstream applications, such non-core elements of a frame are extracted and transformed into stand-alone sentences in our proposed TS approach.

To sum up, our TS approach aims to split sentences that present a complex
structure into its basic semantic building blocks in the form of events. Complex
sentences that contain several events are decomposed into separate frames, while
non-core elements of a frame 
are extracted and
transformed into stand-alone sentences. Hence, from a semantic point of view, a
minimal proposition can be seen as an utterance expressing a single event consisting
of a predicate and its core arguments.

\section{Execution Order of the Patterns}
\label{app:application_order}

The execution order of the transformation patterns was determined by examining which sequence achieved the best simplification results in a manual qualitative analysis that was conducted on a random sample of 300 sentences from Wikipedia. 

The execution order is given in Table \ref{tab:execution_order}. The 35 rules could be grouped into 17 classes.

\begin{table*}[!htb]
\centering
\footnotesize
\begin{tabular}{c | p{8cm}}
\toprule
    \textsc{Order} & \textsc{Rule Group} \\ \hline
     1 & coordinate clauses \\ 
     2 & non-restrictive relative clauses \\ 
     3 & appositive phrases \\ 
     4 & preposed adverbial clauses and participial phrases \\ 
     5 & coordinate verb phrases \\ 
     6 & postposed adverbial clauses and participial phrases \\ 
     7 & reported speech with preposed attribution \\ 
     8 & postposed adverbial clauses \\ 
     9 & reported speech with postposed attribution \\ 
     10 & embedded participial phrases \\ 
     11 & restrictive relative clauses \\ 
     12 & prepositional phrases that act as complements of verb phrases \\ 
     13 & postposed participial phrases \\ 
     14 & adjectival/adverbial phrases \\ 
     15 & lead noun phrases \\ 
     16 & prepositional phrases that are offset by commas \\ 
     17 & lists of noun phrases \\
     \bottomrule
\end{tabular}

\caption{Execution order of the patterns.}
\label{tab:execution_order}
\end{table*}

The execution order of the transformation patterns was balanced for various criteria, including the frequency of the rules, their complexity and specificity, as well as susceptibility to errors. In general, the least error-prone rules are executed first (e.g., in a manual analysis, we determined that 99.1\% of the coordinate clauses (1) were correctly split). During the transformation process, we then work our way to the more error-prone ones, such as the rules for decomposing lists of noun phrases (17), which can easily be confused with appositive phrases. Beyond that, the patterns become more and more complex towards the end of the list, since we need to check for special cases and possibilities of confusion, e.g. (12) and (17). It also becomes clear that we first operate on coarse-grained level, dividing up clauses (1-11), before we go down to the phrasal level (10-17), resulting in much more fine-grained splits. Moreover, specific rules are carried out early in the process, such as the rules for disembedding relative clauses (lexicalized on the relative pronouns, 2) and appositive phrases (based on named entities, 3). More general rules, on the other hand, tend to be executed later. Finally, there is a rough orientation on the frequency of the rules, starting with the ones that are triggered more often, and working towards the ones that occur less frequently in the transformation process.

To sum up, the execution order of the transformation patterns is based on the following criteria, which have been balanced against each other:

\begin{compactitem}
\item[(1)] \textbf{susceptibility to errors}: from less to more error-prone
\item[(2)] \textbf{granularity}: from clausal to phrasal units
    \item[(3)] \textbf{complexity}: from simple to complex
    \item[(4)] \textbf{specificity}: from specific to general
    \item[(5)] \textbf{frequency}: from frequent to infrequent
\end{compactitem}

\section{Constituency Type Classification}
\label{app:constituency}

In RST, each text span is specified as either a nucleus or a satellite. The \textit{nucleus} span embodies the central piece of information and is comparable to what we denote a core sentence, whereas the role of the \textit{satellite} is to further specify the nucleus, corresponding to a context sentence in our case.

\section{Mapping of Cue Phrases to Rhetorical Relations}
\label{appendix:mapping_cue_phrases}

Table \ref{app:mapping_cue_phrases} lists the full set of cue phrases that serve as lexical features for the identification of rhetorical relations 
when establishing the semantic hierarchy between a pair of split sentences. 
 It further shows to which rhetorical relation each of them is mapped.

\begin{table*}[!htb]
\centering
\footnotesize
\begin{tabular}{p{2cm} | p{13cm}}
\toprule
    \textsc{Rhetorical Relation} & \textsc{Cue Phrases} \\ \hline\hline
     \textbf{Contrast} & although, but, but now, despite, even though, even when, except when, however, instead, rather, still, though, thus, until recently, while, yet \\ \hline
     \textbf{List} & and, in addition, in addition to, moreover \\ \hline
     \textbf{Disjunction} & or \\ \hline
     \textbf{Cause} & largely because, because, since \\ \hline
     \textbf{Result} & as a result, as a result of \\ \hline
     \textbf{Temporal} & after, and after, next, then, before, previously \\ \hline
     \textbf{Background} & as, now, once, when, with, without \\ \hline
     \textbf{Condition} & if, in case, unless, until \\ \hline
     \textbf{Elaboration} & more provocatively, even before, for example, further, recently, since, since now, so, so far, where, whereby, whether \\ \hline
     \textbf{Explanation} & simply because, because of, indeed, so, so that \\
     \bottomrule
\end{tabular}

\caption{Mapping of cue phrases to rhetorical relations.}
\label{app:mapping_cue_phrases}
\end{table*}

In addition, \textit{Spatial} relations are identified on the basis of named entities. \textit{Attribution} relations are detected using a pre-defined list of verbs of reported speech and cognition \cite{carlson2001discourse}.

\section{Merits for Downstream Applications}
\label{app:merits}


\paragraph{Coherence} 
Previous work in the area of Open IE has mainly focused on the extraction of
isolated relational tuples, ignoring the cohesive nature of texts where important
contextual information is spread across clauses or sentences. Consequently, state-of-the-art Open IE approaches are prone to generating a loose arrangement of tuples that lack the expressiveness needed to infer the true meaning of complex assertions \cite{niklaus-etal-2018-survey}.

\begin{figure*}[!htb]
  \flushleft
  \scriptsize
  
  \begin{verbatim}
  RnnOIE (stand-alone):
  (1) (A fluoroscopic study;                     known;                 as an upper gastrointestinal series)
  (2) (caution with non water soluble contrast;  is;                    mandatory as the usage of barium)
  (3) (as the usage;                             of barium can impede;  surgical revision and lead)
  (4) ( ;                                        to increased;          post operative complications)}
  
  RnnOIE (with pre-processing):
  (5)  #1    0    (A fluoroscopic study;              is;   typically, the next step in management)
  (5a)                L:ELABORATION   #2
  (5b)                L:CONTRAST      #3
  (6)  #2    1    (This;    fluoroscopic study is known;    as an upper gastrointestinal series)
  (7)  #3    0    (Caution with non water soluble;    is;   mandatory)
  (7a)                L:CONTRAST      #1
  (7b)                L:CONDITION     #7
  (7c)                L:BACKGROUND    #4
  (7d)                L:BACKGROUND    #5
  (7e)                L:BACKGROUND    #6
  (8)  #4    1    (The usage of barium;    can impede;      surgical revision)
  (8a)                L:LIST          #5
  (8b)                L:LIST          #6
  (9)  #5    1    (The usage of barium;    can lead;        to increased post operative complications)
  (9a)                L:LIST          #4
  (9b)                L:LIST          #6
  (10) #6    1    (The usage of barium;    to increased;    post operative complications)
  (10a)               L:LIST          #4
  (10b)               L:LIST          #5
  (11) #7    1    (Volvulus;               is suspected;    )
  \end{verbatim}
  
  \caption{Comparison of the relational tuples extracted by RnnOIE \cite{stanovsky2018supervised} \textit{with} and \textit{without} using our context-preserving TS approach as a preprocessing step.}
  \label{fig:ComparativeAnalysisSystems_supervisedOIE}
\end{figure*}

Our approach allows existing Open IE systems to enrich their output with semantic information.
By leveraging the semantic hierarchy of minimal propositions generated by our discourse-aware TS approach,
they are able to extract semantically typed relational tuples from complex source
sentences. Thus, the shallow semantic representation of state-of-the-art Open IE approaches in the form of disconnected predicate-argument structures is extended, capturing intra-sentential rhetorical structures and
hierarchical relationships between the relational tuples. In that way, the semantic
context of the extracted tuples is preserved, allowing for a proper interpretation of
complex assertions. See Figure \ref{fig:ComparativeAnalysisSystems_supervisedOIE} for an illustrative example. 

\section{Evaluation}
\label{app:evaluation}

While the goal of our TS approach is to generate well-formed syntactically simplified sentences, the EDUs in the RST-DT are copied verbatim from the source, 
 resulting in an output of varied length that is usually not grammatically sound. Moreover, in many cases, the EDUs mix multiple semantic units, whereas our approach aims to split the input into atomic components, with each of them expressing a coherent and indivisible proposition.


\begin{figure*}[!htb]
    \centering
    \includegraphics[width=\textwidth]{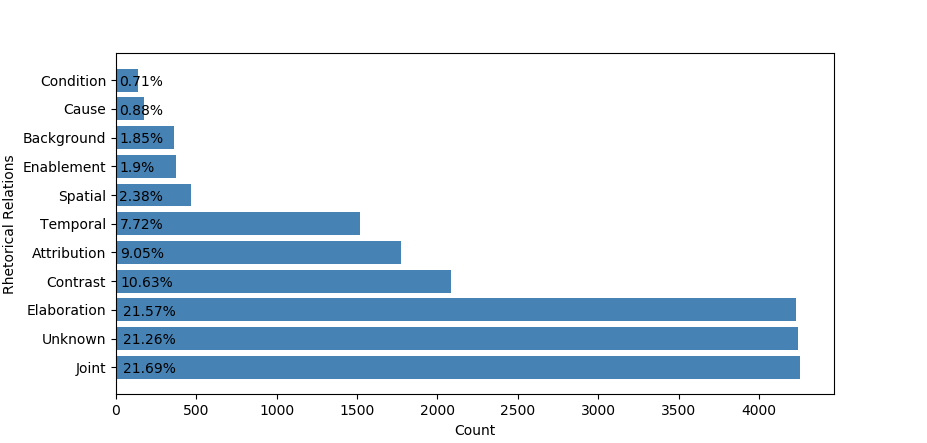}
    \caption{Distribution of the rhetorical relations.} 
    \label{fig:distribution_rhetorical_relations_dissim}
\end{figure*}

\section{Discourse Parsing}
\label{app:related_work}
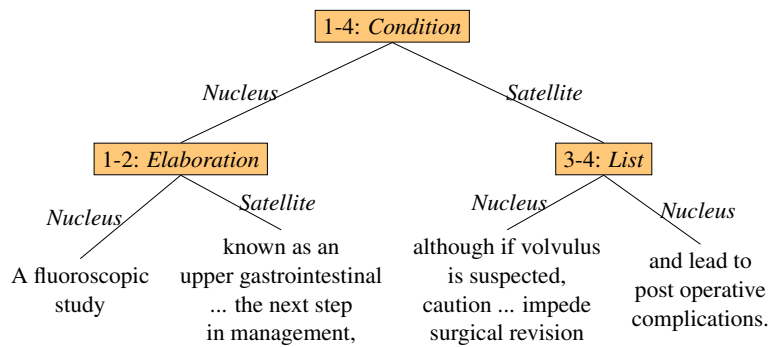
\begin{figure*}[!htb]
\centering
\begin{tikzpicture}[scale=0.8, level distance=2.2cm, sibling distance=0.25cm, every tree node/.style={align=center, transform shape}]
\Tree [
      .\node [style={draw,rectangle}, fill={rgb:orange,1;yellow,2;pink,5}] {1-4: \textit{Condition}};
            \edge node[midway, left] {\textit{Nucleus}}; [.\node [style={draw,rectangle}, fill={rgb:orange,1;yellow,2;pink,5}] {1-2: \textit{Elaboration}};
             \edge node[midway, left] {\textit{Nucleus}}; [.\node(a){A fluoroscopic\\ study};]
             \edge node[midway, right] {\textit{Satellite}}; [.\node(b){known as an\\ upper gastrointestinal \\ ... the next step \\ in management,};]
             ]
             \edge node[midway, right] {\textit{Satellite}}; [.\node [style={draw,rectangle}, fill={rgb:orange,1;yellow,2;pink,5}] {3-4: \textit{List}};
             \edge node[midway, left] {\textit{Nucleus}}; [.\node(a){although if volvulus\\ is suspected,\\ caution ... impede \\ surgical revision};]
             \edge node[midway, right] {\textit{Nucleus}}; [.\node(b){and lead to \\ post operative \\ complications.};]
             ]
        ]
]
\end{tikzpicture}
\caption{Rhetorical structure tree of our example sentence, generated using the RST parser proposed in \newcite{ji-eisenstein-2014-representation}. The leaves correspond to \textit{EDUs}, while each node is characterized by its \textit{nuclearity} and a \textit{rhetorical relation} between adjacent text spans.}
\label{exampleRST}
\end{figure*}

The syntactic analysis we propose for establishing the semantic hierarchy between the decomposed spans is bound to the RST discourse markers. However, it is not possible to simply use an RST parser for this task. As illustrated in Figure \ref{exampleRST}, such a parser does not return grammatically sound sentences. Instead, it segments the input into basic textual units, so-called elementary discourse units (EDUs), which are copied verbatim from the source. In order to reconstruct proper sentences, rephrasing is required. For this purpose, amongst others, referring expressions have to be identified, and phrases have to be rearranged and inflected. Moreover, the textual units resulting from the segmentation process are too coarse-grained for our purpose, since RST parsers mostly operate on clausal level. The goal of our approach, though, is to split the input into minimal semantic units, which requires to go down to the phrasal level in order to produce a much more fine-grained output in the form of minimal propositions.

\end{document}